\documentclass[11pt,a4paper]{article}
\usepackage[hyperref]{ranlp2023}
\usepackage{times}
\usepackage{latexsym}
\usepackage{subcaption}
\usepackage{graphicx}
\usepackage{mathtools}
\usepackage{hyperref}

\usepackage[colorinlistoftodos]{todonotes}

\usepackage{microtype}
\usepackage{multirow}
\usepackage{rotating}
\aclfinalcopy 

\title{ARC-NLP at Multimodal Hate Speech Event Detection 2023: Multimodal Methods Boosted by Ensemble Learning, Syntactical and Entity Features}

\author{Umitcan Sahin, Izzet Emre Kucukkaya, Oguzhan Ozcelik, Cagri Toraman \\
  Aselsan Research Center, Ankara, Turkiye \\
  \texttt{\{ucsahin, ekucukkaya, ogozcelik, ctoraman\}@aselsan.com.tr} 
}
\date{}

\begin{document}
\maketitle

\begin{abstract}
Text-embedded images can serve as a means of spreading hate speech, propaganda, and extremist beliefs.
Throughout the Russia-Ukraine war, both opposing factions heavily relied on text-embedded images as a vehicle for spreading propaganda and hate speech. 
Ensuring the effective detection of hate speech and propaganda is of utmost importance to mitigate the negative effect of hate speech dissemination. 
In this paper, we outline our methodologies for two subtasks of Multimodal Hate Speech Event Detection 2023. For the first subtask, hate speech detection, we utilize multimodal deep learning models boosted by ensemble learning and syntactical text attributes. For the second subtask, target detection, we employ multimodal deep learning models boosted by named entity features. 
Through experimentation, we demonstrate the superior performance of our models compared to all textual, visual, and text-visual baselines employed in multimodal hate speech detection. Furthermore, our models achieve the first place in both subtasks on the final leaderboard of the shared task.
\end{abstract}

\section{Introduction}
The Russia-Ukraine War has been a long and bitter conflict that has caused a lot of division and tension among people. Unfortunately, hate speech has played a big role in this war, spreading negativity, fueling hatred, and making the situation even more volatile. It is important to find ways to detect and combat hate speech in order to promote unity and peace.

Deep learning models are increasingly being employed in multimodal hate speech detection \cite{Dataset:2023, Task:2023, Boishakhi:2021, Gomez:2020, Yang:2019, Perifanos:2021, Rana:2022, Vijayaraghavan:2021, Sabat:2019, madukwe:2020, Kiela:2020}. These models leverage the power of neural networks to process and analyze complex data consisting of text, images, and videos, allowing them to capture the nuances and context of online content. By combining various modalities, such as textual and visual contents, these models can better understand the overall meaning and intent behind the shared information. They learn from large amounts of labeled data, enabling them to identify patterns and distinguish between genuine information and harmful content, including hate speech and misinformation \cite{Toraman:2022}. With their ability to integrate multiple modalities, deep learning models are playing a vital role in combating online abuse, fostering safer digital environments, and promoting responsible information dissemination.

This study addresses the challenge of combating hate speech using multiple modalities, specifically focusing on the shared task of Multimodal Hate Speech Event Detection at CASE 2023. 
In the shared task, Subtask A requires determining whether a text-embedded image contains hate speech. To address this, we propose a novel ensemble model that merges predictions from a multimodal deep learning model and multiple text-based tabular models which are trained with various syntactical features.
On the other hand, for Subtask B, the goal is to identify the target of hate speech in a text-embedded image and classify it into the categories of ``Individual", ``Community", or ``Organization". To tackle this challenge, we introduce a novel multimodal deep learning model. We train a multimodal deep learning model and then combine its embeddings with named entity features, which are then used as input to train a new fusion model. Through experimentation, we show that our proposed models achieve superior classification performance compared to the multimodal hate speech detection baselines. Notably, our proposed models achieve the highest rank on the final leaderboard for both subtasks in the shared task. 

\section{Dataset \& Task}
\begin{table}
\centering
\resizebox{\columnwidth}{!}{%
\begin{tabular}{cclrrr}
\hline \multirow{2}{*}{\textbf{Subtask}} & \multirow{2}{*}{\textbf{Problem}} & \multirow{2}{*}{\textbf{Labels}} & \multicolumn{3}{c}{\textbf{\#Text-embedded Images}} \\ 
& & & \multicolumn{1}{r}{Train} & \multicolumn{1}{r}{Eval} & \multicolumn{1}{r}{Test} \\
\hline
\multirow{2}{*}{A} & Hate & Hate & 1,942 & 243 & \multirow{2}{*}{443}  \\
& Speech & Non-Hate & 1,658 & 200 &  \\ 
\hline
\multirow{3}{*}{B} & \multirow{3}{*}{Target} & Individual & 823 & 102 & \multirow{3}{*}{242} \\
& & Community & 335 & 40 & \\
& & Organization & 784 & 102 & \\
\hline
\end{tabular}}
\caption{ The dataset for the shared task on Multimodal Hate Speech Event Detection at CASE 2023. The numbers of text-embedded images in the train, evaluation and test sets for both Subtask A and B are given. The labels of the test set examples are not shared.}
\label{tab:dataset}
\end{table}

The shared task on Multimodal Hate Speech Event Detection at CASE 2023\footnote{https://github.com/therealthapa/case2023\_task4} consists of two distinct subtasks: Subtask A and B. The details of each subtask are presented in Table \ref{tab:dataset} along with the number of text-embedded images in the training, evaluation and test sets. It is important to note that the labels of the test set examples are not disclosed to the participants during the shared task. These labels are reserved for calculating the final prediction performance, which determines the leaderboard rankings upon completion of the shared task. Furthermore, text within the images are extracted using OCR with Google Vision API\footnote{https://cloud.google.com/vision/docs/ocr}.   

\subsection{Subtask A: Hate Speech Detection}
In Subtask A, it is aimed to determine the presence or absence of hate speech within text-embedded images \cite{Task:2023}. The dataset specifically designed for this subtask includes annotated examples that indicate the existence of hate speech \cite{Dataset:2023}. The dataset features two distinct labels: ``Hate Speech" and ``No Hate Speech". 

\subsection{Subtask B: Target Detection}
Subtask B aims to identify the targets of hate speech within a given hateful text-embedded image \cite{Task:2023}. The dataset provided for this subtask includes labels categorizing the hate speech targets into ``Individual", ``Community", and ``Organization" \cite{Dataset:2023}.

\section{Methodology}
\label{sec:methods}
In this section, we describe our proposed models for Subtask A and B of the shared task, respectively. 

\begin{figure*}
  \begin{subfigure}[b]{\columnwidth}
    \includegraphics[width=\linewidth]{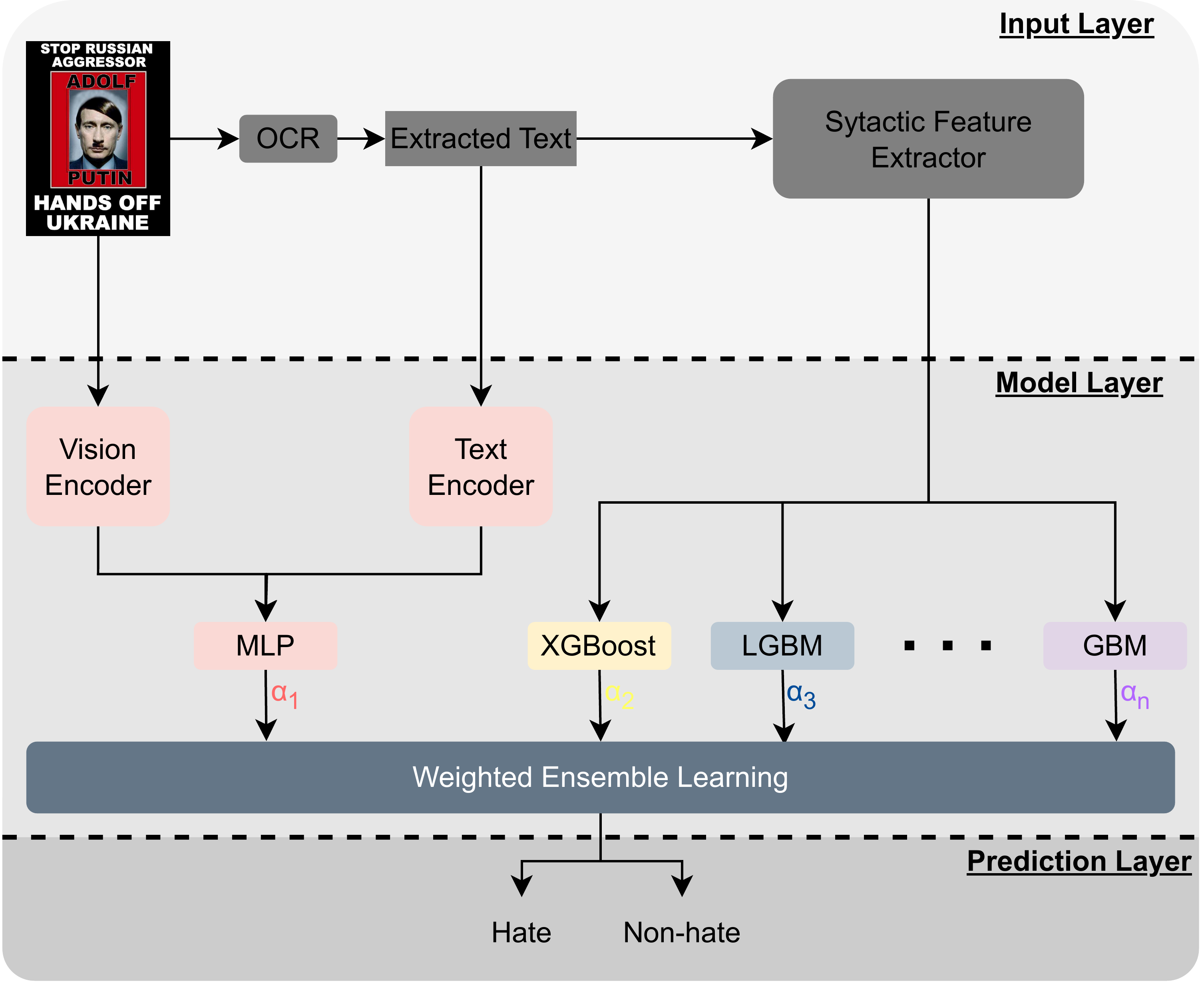}
    \caption{}
    \label{fig:subtask-a-overlay}
  \end{subfigure}
  \hfill 
  \begin{subfigure}[b]{\columnwidth}
    \includegraphics[width=\linewidth]{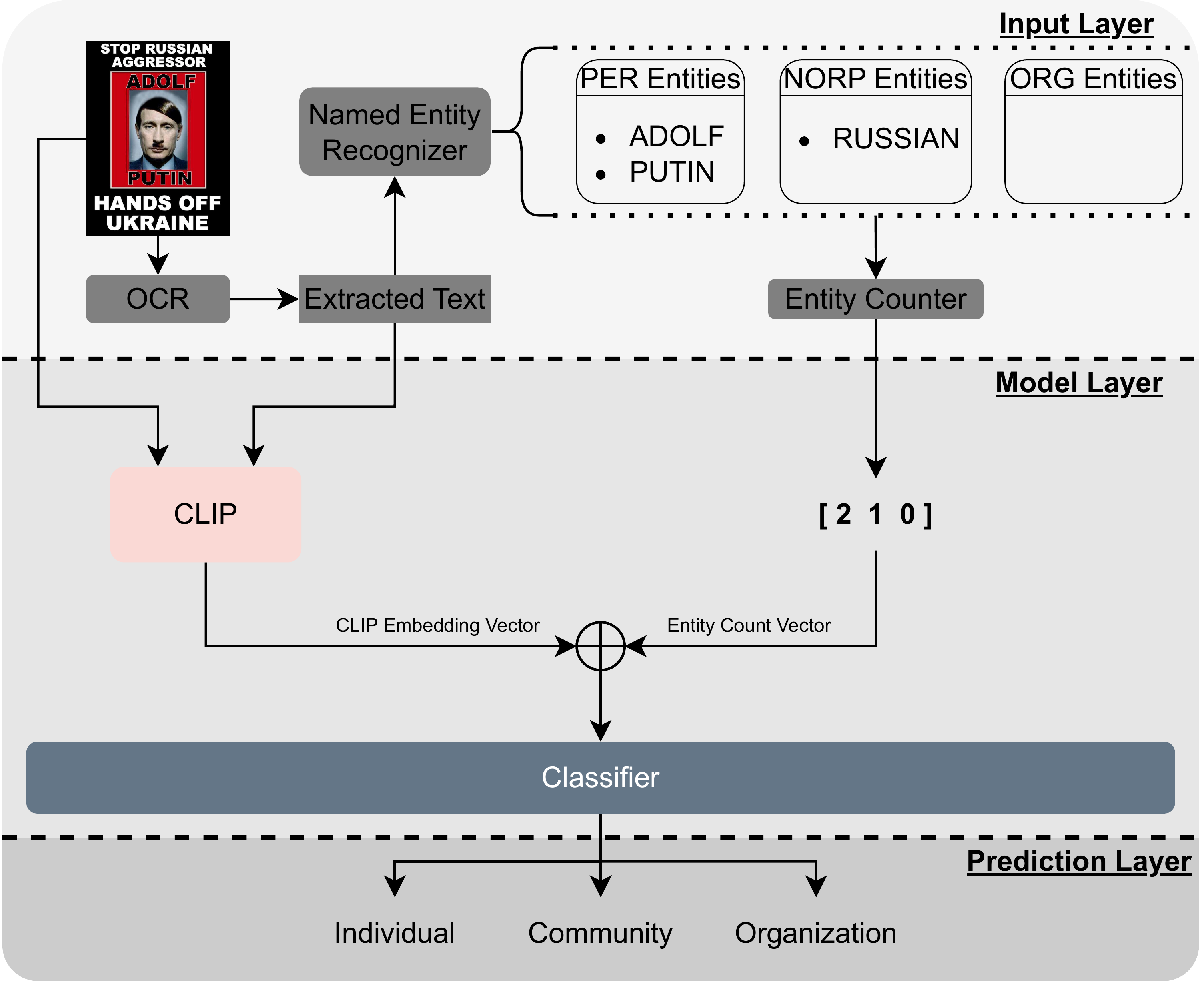}
    \caption{}
    \label{fig:subtask-b-overlay}
  \end{subfigure}
  \caption{The high-level illustrations of our models for (a) Subtask A and (b) Subtask B. Each model consists of three stages, which are the Input, Model, and Prediction layers. The Input layer describes the processes of text and syntactic feature extraction, and entity recognition. In the Model layer, we indicate the training procedures. For the joint learning of the models we represent the same colored blocks. For instance, in (a) Vision and Text encoder, and MLP is jointly trained, while XGBoost, LGBM, and GBM have independent training procedures. The last layer, i.e., Prediction, shows the classified labels for each model. }
  \label{fig:subtask-a-and-b}
\end{figure*}
\subsection{Proposed model for Subtask A: Ensemble of multimodal deep learning and text-based tabular models}
\label{sec:subtask-A}
The process of identifying hate speech within an image and its OCR-generated text can be approached using various methods, including relying solely on image-based or text-based models. However, in our approach, we adopt a multimodal approach to leverage the full knowledge present in the dataset. We employ both textual and visual features to train our deep learning models, aiming to capture a comprehensive understanding of the data. Additionally, we incorporate various syntactical features into our model. For this, we construct a $33$-dimensional syntactical feature vector as shown in Table \ref{tab:syntactical}.

\begin{table}[]
    \centering
    \begin{tabular}{p{4cm} c}
    \textbf{Feature} & \textbf{Count} \\
    \hline
    Word counts & 1 \\
    Character counts & 1\\
    Capital ratio & 1\\
    Digit ratio & 1\\
    Special character ratio &1\\
    White space ratio & 1\\
    Symbol (!, ?, @, \%, *, \$, \&, \#, ., :, /, -, =) ratios & 13\\
    Symbol counts & 13 \\
    Lowercase ratio & 1 \\
    \hline
    \end{tabular}
    \caption{Syntactical features used in our proposed model for Subtask A.}
    \label{tab:syntactical}
\end{table}

Furthermore, we also use the Bag-of-words (BoWs) method to extract n-grams ($n \in \{1,2,3\}$) from text and use them as additional features. This choice is motivated by our observation that the BoW method has competitive performance in hate speech detection and these features might possibly serve as indicators of hate speech, independent of the overall meaning conveyed by the text and image \cite{Toraman:2022b}.

As illustrated in Figure \ref{fig:subtask-a-overlay}, our methodology begins by combining a text encoder with a vision encoder model via a multi-layer perceptron (MLP) module. This multimodal structure is initially trained on the entire training set using a linear classifier layer with the cross-entropy loss function. We select the best-performing model based on the accuracy metric across multiple training epochs using the evaluation set. Subsequently, we extract the aforementioned syntactical and BoW features from the text, which are then used to train tabular learning models (i.e., classifiers), including LightGBMXT, LightGBMLarge, LightGBM \cite{Ke:2017}, CatBoost \cite{Prokhorenkova:2018}, and XGBoost \cite{Chen:2016}. We then combine these models to maximize the utilization of available information. To accomplish this, we adopt an ensemble approach similar to our previous work in CASE2022 \citep{Hurriyetoglu:2022, Sahin:2022}. However, this time we utilize a weighted ensembler which assigns adaptive weights to each model and generates final predictions based on these weights. The weight assignment is determined during the training phase and optimized with respect to the validation accuracy computed on the evaluation set of Subtask A.

\subsection{Proposed model for Subtask B: Combining multimodal deep learning with named entity recognition}
In our proposed model for Subtask B, instead of using syntactical features, we employ named entities which are extracted from the text. Named entity recognition (NER) aims to extract important information from unstructured text \cite{Ozcelik:2022} and can be used as a supportive feature to improve the classification performance of a deep learning model. Therefore, we obtain named entities for the unstructured texts extracted from the text-embedded images using the spaCy library \cite{Honnibal:2017}. SpaCy is an open source NLP library including several tasks such as Part-of-Speech (POS) tagging and NER. We use the English pretrained large NER model\footnote{\href{https://github.com/explosion/spacy-models/releases/tag/en_core_web_lg-3.6.0}{en\_core\_web\_lg-3.6.0}} as a named entity recognizer (see Figure \ref{fig:subtask-b-overlay}). The motivation behind using this model is that it contains individual, community, and organization named entity classes, which are directly related to the prediction classes of Subtask B. Therefore, we only extract \texttt{PER}, \texttt{NORP}, and \texttt{ORG} entities as shown in Figure \ref{fig:subtask-b-overlay}. The \texttt{PER} entities include people or fictional character names. The \texttt{NORP} entities represent nationalities or religious and political groups (e.g., communities). Finally, the \texttt{ORG} entities are referred to organization names, such as NATO. After we obtain the aforementioned entities from the extracted texts of the images, we generate a feature vector, consisting of the counts of each entity. For instance, from Figure \ref{fig:subtask-b-overlay}, we represent the vector for the extracted text ``STOP RUSSIAN AGRESSOR ADOLF PUTIN HANDS OFF UKRAINE'' as $\begin{bmatrix} 2 & 1 & 0 \end{bmatrix}$ since two (i.e., \textit{Putin}, \textit{Adolf}), one (i.e., \textit{Russian}), and no entities are obtained for \texttt{PER}, \texttt{NORP}, and \texttt{ORG} classes, respectively.

Figure \ref{fig:subtask-b-overlay} shows the overall structure of our proposed model for Subtask B. Using the text-embedded images and the extracted OCR text from these images in the training set, we first fine-tune a Contrastive Language-Image Pre-Training (CLIP) model, which is a multimodal deep learning model that is pre-trained on a variety of (image, text) pairs \cite{Radford:2021}. Following the completion of the CLIP training, we proceed to extract the embedding vector for each (image, text) pair in the training set of Subtask B. These embedding vectors and the entity count vector are then concatenated together to create a novel fusion vector. This newly formed vector serves as the input for training multiple tabular learning models (i.e., classifiers), including LightGBMLarge, LightGBM, and XGBoost. The classifier that achieves the highest validation accuracy score on the evaluation set of Subtask B is then selected to generate final predictions.

\section{Results \& Discussion}
\subsection{Baselines}
We employ the AutoGluon framework \cite{Erickson:2020} for the implementation of our proposed models and the baselines for multimodal hate speech detection. AutoGluon is an AutoML toolkit and provides a comprehensive environment for multimodal training. We use the following hyperparameter setting for the training of all models: The learning rate is set to 1e-4, learning rate decay is set to 0.9, learning rate scheduler is cosine decay, maximum number of epochs is 10, warm-up step is 0.1, per GPU batch size is 8. During the training phase of our models and the baselines, we utilize four NVIDIA A4000 GPUs. We categorize the baselines into four categories: Tabular, Textual, Visual or Multimodal, which are explained below.

\subsubsection{Tabular Baselines}
For the tabular baseline models, we construct syntactic features derived from the textual data. These features, which are shown in Table \ref{tab:syntactical}, and BoW features (i.e., n-grams with $n \in \{1,2,3\}$) are employed to train classifiers including LightGBMXT, LightGBMLarge, LightGBM, CatBoost, and XGBoost. We use the AutoGluon implementation of the classifiers with default parameters.

\subsubsection{Textual Baselines}
For the text-only baseline models, we use the following transformer-based language models: BERT (BERT-base-cased\footnote{https://huggingface.co/bert-base-cased}) \cite{Devlin:2019}, RoBERTa (RoBERTa-base\footnote{https://huggingface.co/roberta-base}) \cite{Liu:2019}, DeBERTa-v3 (DeBERTa-v3-base\footnote{https://huggingface.co/microsoft/deberta-v3-base}) \cite{He:2023}, and ELECTRA (ELECTRA-base-discriminator\footnote{https://huggingface.co/google/electra-base-discriminator}). We use the AutoGluon implementation of the models with a maximum token size of 512 and padding the rest.

\subsubsection{Visual Baselines}
For the image-only baseline models, we employ the following transformer-based encoders: Swin (swin-base-patch4-window7-224\footnote{https://huggingface.co/microsoft/swin-base-patch4-window7-224-in22k}), CoAtNet-v3 (coatnet-v3-rw-224-sw\_in12k\footnote{https://huggingface.co/timm/coatnet\_3\_rw\_224.sw\_in12k}) \cite{Dai:2021}, DaViT (davit-base-msft-in1k\footnote{https://huggingface.co/timm/davit\_base.msft\_in1k}) \cite{Ding:2022}, and ViT (vit-base-patch32-224-in21k\footnote{https://huggingface.co/google/vit-base-patch32-224-in21k}) \cite{Dosovitskiy:2021}. We use the AutoGluon implementation of the models with default parameters.

\subsubsection{Multimodal Baselines}
For the multimodal baseline models where both text and images are used in the training process, we combine a textual and a visual baseline model together and jointly train them by using a multi-layer perceptron (MLP) on top of them with a binary cross-entropy loss function. To determine the optimal combination of the models, we select the top-performing text and vision encoders based on their individual performances in terms of the validation accuracy score computed on the evaluation set of the corresponding subtasks. For this, we employ the AutoGluon implementation of the text and vision encoders with a maximum token size of 512 and all other parameters set to their default values. For the classification layer, we use two fully connected linear layers (128 dimensional hidden layer) with a Leaky ReLU activation function between them. Furthermore, we also use the AutoGluon's implementation of the CLIP model as one of the multimodal baselines.  

\begin{table*}
    \centering
    \small
    \begin{tabular}{llcccc}
    \hline
    & \textbf{Model} & \textbf{Precision} & \textbf{Recall} & \textbf{F1} & \textbf{Accuracy} \\
    \hline
    \multirow{5}{*}{\begin{turn}{90}Tabular\end{turn}} & XGBoost & 82.0 & 82.7 & 80.6& 80.6\\
    & LightGBM & 81.2 & 83.5 & 80.3 & 80.4 \\
    & LightGBMLarge & 81.6 & 82.3 & 80.1 & 80.1 \\
    & CatBoost & 79.7 & 82.3 & 78.7 & 78.8 \\
    & LightGBMXT & 78.8 & 81.1 & 77.6 & 77.6 \\
    \hline
    \multirow{4}{*}{\begin{turn}{90}Textual\end{turn}} &  ELECTRA & 82.2 & 89.3 & 83.4 & 83.5\\
    & BERT & 79.4 & 84.4 & 79.4 & 79.4 \\
    & RoBERTa & 84.3 & 81.9 & 81.7 & 81.7\\
    & DeBERTa-v3 & 83.0 & 86.4 & 82.8 & 82.8\\
    \hline
    \multirow{4}{*}{\begin{turn}{90}Visual\end{turn}}& Swin & 74.7 & 84.0 & 75.3 & 75.6\\
    & CoAtNet-v3 & 80.4 & 81.1 & 78.8 & 78.8\\
    & DaViT & 81.5 & 79.2 & 78.1 & 78.1 \\
    & ViT & 79.0 & 77.7 & 76.0 & 76.1\\
    \hline
    \multirow{5}{*}{\begin{turn}{90}Multimodal\end{turn}}& ELECTRA + Swin & 83.3 & 90.1 & 84.5 & 84.6\\
    & DeBERTa-v3 + Swin & 81.8 & 90.9 & 83.8 & 84.0\\
    & ELECTRA + CoAtNet-v3 & 85.4 & 86.4 & 84.4 & 84.4\\
    & DeBERTa-v3 + CoAtNet-v3 & 82.9 & 87.6 & 83.2 & 83.3\\
    & CLIP & 79.9 & 91.8 & 82.6 & 82.8\\
    \hline
    \textit{Ours} & \textbf{ELECTRA + Swin + Tabular} & \textbf{84.1} & \textbf{89.0} & \textbf{84.8}& \textbf{84.9}\\
    \hline
    \end{tabular}
    \caption{\textbf{Subtask A: Hate Speech Detection} evaluation results in terms of binary precision, recall, F1-score, and accuracy metrics. Tabular, textual, visual, and multimodal baselines are implemented using the AutoGluon library \cite{Erickson:2020} and categorized into their respective categories. The model which achieves the highest test scores on the final leaderboard is indicated with a bold font. }
    \label{tab:subtask-a-results}
\end{table*}

\subsubsection{Our Models}
For the implementation of our proposed models for Subtask A and B in Section \ref{sec:methods}, we again employ the AutoGluon library. For Subtask A, we use ELECTRA (ELECTRA-base-discriminator) and Swin (swin-base-patch4 window7-224) as our text and vision encoders, respectively. Using the syntactical and BoW features described in Section \ref{sec:methods}, we train the tabular models LightGBMXT, LightGBMLarge, LightGBM, CatBoost, and XGBoost with default parameters. Additionally, we utilize the weighted ensembler L2, an implementation provided by AutoGluon, to combine the predictions of the individual models and generate final predictions. This weighted ensembling technique assigns weights to each model, taking into account their respective classification performance on the evaluation set of Subtask A. 

Furthermore, for Subtask B, we use the the multimodal baseline CLIP model and combine its embedding vector with NER features as described in Section \ref{sec:methods}. With the combined features, we train a LightGBMlarge classifier with default parameters to produce final predictions. 

\begin{table*}
    \centering
    \small
    \begin{tabular}{llcccc}
    \hline
    & \textbf{Model} & \textbf{Precision} & \textbf{Recall} & \textbf{F1} & \textbf{Accuracy} \\
    \hline
    \multirow{3}{*}{\begin{turn}{90}Tabular\end{turn}} & XGBoost & 65.2 & 64.1 & 63.4 & 65.2 \\
    & LightGBM & 68.0 & 67.3 & 66.6 & 68.0 \\
    & LightGBMLarge & 68.8 & 68.3 & 67.4 & 68.8 \\
    \hline
    \multirow{4}{*}{\begin{turn}{90}Textual\end{turn}} & ELECTRA & 66.0 &	65.6 & 65.7 &	66.0 \\
    & BERT & 66.0 & 64.7 & 64.7 & 66.0 \\
    & RoBERTa & 71.7 &	71.4 &	71.4 &	71.7 \\
    & DeBERTa-v3 & 68.8 & 67.1 & 66.2 & 68.8 \\
    \hline
    \multirow{4}{*}{\begin{turn}{90}Visual\end{turn}}& Swin & 51.3 & 54.5 &	52.0 & 54.5 \\
    & CoAtNet-v3 & 49.5 & 50.8 & 49.9 &	50.8 \\
    & DaViT & 47.9 & 51.6 &	48.5 & 51.6 \\
    & ViT & 42.2 & 45.1 & 42.3 & 45.1 \\
    \hline
    \multirow{5}{*}{\begin{turn}{90}Multimodal\end{turn}} & RoBERTa + CoAtNet-v3 & 68.5 & 69.6 & 68.4 & 69.6 \\
    & DeBERTa-v3 + CoAtNet-v3 & 63.8 & 63.6 & 62.6 & 63.6 \\
    & RoBERTa + Swin & 72.7 & 73.8 & 72.6 & 73.8 \\
    & DeBERTa-v3 + Swin & 66.2 & 66.0 & 65.0 &	66.0 \\
    & CLIP & 74.2 & 76.8 & 75.4 & 76.8 \\
    \hline
    \textit{Ours} & \textbf{CLIP + NER} & \textbf{80.5} & \textbf{80.3} & \textbf{79.7} & \textbf{80.3} \\
    \hline
    \end{tabular}
    \caption{\textbf{Subtask B: Target Detection} evaluation results in terms of weighted precision, recall, F1-score, and multi-class accuracy metrics. Tabular, textual, visual, and multimodal baselines are implemented using the AutoGluon library \cite{Erickson:2020} and categorized into their respective categories. The model which achieves the highest test scores on the final leaderboard is indicated with a bold font. }
    \label{tab:subtask-b-results}
\end{table*}

\subsection{Evaluation Results}
Table \ref{tab:subtask-a-results} and \ref{tab:subtask-b-results} show the classification performance metrics of our models and the baselines computed on the evaluation sets of Subtask A and B, respectively. \textit{Precision}, \textit{Recall}, \textit{F1}, and \textit{Accuracy} metrics are used for measuring the classification performance on the shared task of Multimodal Hate Speech Event Detection at CASE 2023\footnote{https://codalab.lisn.upsaclay.fr/competitions/13087\#results}.

The results in Table \ref{tab:subtask-a-results} and \ref{tab:subtask-b-results} clearly show that our proposed models, along with ensemble learning and using syntactical features for Subtask A and NER features for Subtask B, perform much better than all other methods, including the tabular, textual, visual, and multimodal baselines, for detecting hate speech in a multimodal setting. These results demonstrate that including different text-based features in our models improves their performance significantly, allowing us to make better use of the information in the dataset. This emphasizes the importance of using various textual attributes to enhance the overall effectiveness of the models.   

In our experiments, we observe that textual methods trained with the extracted OCR text from the text-embedded images outperform visual methods trained solely on images. Additionally, the tabular models, which are trained with syntactical and BoW features (i.e., n-grams, $n \in \{1,2,3\}$), achieve results comparable to the text-based methods. This once again demonstrates the effectiveness of these features in multimodal hate speech detection.

Furthermore, multimodal approaches that combine multiple modalities, such as image and text, effectively leverage both textual and visual information, resulting in significantly more powerful deep learning models. This integration of different modalities enhances the overall performance of the models in the process.

Finally, introducing a named entity recognition (NER) system capable of extracting key elements from unstructured text, like person names, organizations, and locations, proves particularly effective in identifying targets of hate speech (e.g., individuals, communities, and organizations) within a given text. By incorporating NER features into our model for Subtask B, we are able to further enhance the classification performance of the multimodal methods. This improvement is clearly demonstrated by the classification performance of our proposed model, as illustrated in Table \ref{tab:subtask-b-results}.

\begin{table}
\centering
\resizebox{\columnwidth}{!}{%
\begin{tabular}{lrrrr}
\hline \textbf{Team Name} & \textbf{Recall} & \textbf{Precision} & \textbf{F1} & \textbf{Accuracy} \\ \hline
\textbf{ARC-NLP} & \textbf{85.67} & \textbf{85.63} & 	\textbf{85.65} & 	\textbf{85.78} \\
bayesiano98 & 85.61 & 	85.28 & 	85.28 & 	85.33  \\
IIC\_Team & 85.08 & 	84.76 & 	84.63 & 	84.65 \\
DeepBlueAI & 83.56 & 	83.35 & 	83.42 & 	83.52  \\
CSECU-DSG & 82.52 & 	82.44 & 	82.48 & 	82.62\\
Ometeotl &  	81.21 & 	80.94 & 	80.97 & 	81.04  \\
Avanthika & 78.78 & 	78.81 & 	78.80 & 	79.01  \\
Sarika22 & 78.06 & 	78.49 & 	78.21 & 	78.56  \\
rabindra.nath &  77.68 & 	78.42 & 	77.88 & 	78.33  \\
md\_kashif\_20 & 72.70 & 	73.72 & 	72.87 & 	73.59  \\
GT & 52.19 & 	52.19 & 	52.19 & 	52.60  \\
Team +1 & 49.38 & 	49.39 & 	49.36 & 	49.66  \\
ML\_Ensemblers & 53.34 & 	72.40 & 42.94 & 57.79  \\
\hline
\end{tabular} }
\caption{The leaderboard results of \textbf{Subtask A: Hate Speech Detection}. Our team name is \textbf{ARC-NLP}. The teams are ranked by the F1 score. Our solution is ranked first in terms of all classification metrics. }
\label{tab:leaderboard-a}
\end{table}

\begin{table}
\centering
\resizebox{\columnwidth}{!}{%
\begin{tabular}{lrrrr}
\hline \textbf{Team Name} & \textbf{Recall} & \textbf{Precision} & \textbf{F1} & \textbf{Accuracy} \\ \hline
\textbf{ARC-NLP} & \textbf{76.36} & \textbf{76.37} & \textbf{76.34} & \textbf{79.34} \\
bayesiano98 & 73.30 & 75.54 & 74.10 & 77.27 \\
IIC\_Team & 68.94 & 71.05 & 69.73 & 72.31 \\
Sarika22 & 67.77 & 	68.41 & 68.05 & 71.49 \\
CSECU-DSG & 65.25 & 65.75 & 65.30 & 69.01 \\
DeepBlueAI & 64.62 & 66.48 & 65.25 & 69.83 \\
Ometeotl &  56.48 & 67.93 & 56.77 & 64.05 \\
Avanthika & 53.84 & 70.13 & 52.58 & 64.05 \\
ML\_Ensemblers & 44.44 & 48.88 & 43.32 & 52.89 \\
Team +1 & 34.42 & 	35.59 & 33.42 & 35.12 \\
\hline
\end{tabular} }
\caption{The leaderboard results of \textbf{Subtask B: Target Detection}. Our team name is \textbf{ARC-NLP}. The teams are ranked by the F1 score. Our solution is ranked first in terms of all classification metrics. }
\label{tab:leaderboard-b}
\end{table}

\subsection{Leaderboard Results}
During the test phase of the shared task, we submitted our models to be evaluated on the test sets of both Subtask A and Subtask B. The test results have been presented in Table \ref{tab:leaderboard-a} and Table \ref{tab:leaderboard-b}, respectively.

Our model, \emph{ELECTRA+Swin+Tabular}, achieved the top rank among 13 participating teams in Subtask A, excelling in all classification metrics within the test results. Similarly, our model, \emph{CLIP+NER}, secured the first position among 10 participating teams in Subtask B, performing exceptionally well across all classification metrics.

\section{Conclusion}
In conclusion, the utilization of text-embedded images on social media has become a common means of expressing opinions and emotions. However, it has also been exploited to spread hate speech, propaganda, and extremist ideologies, as witnessed during the Russia-Ukraine war. Detecting and addressing such instances are crucial, particularly in times of ongoing conflict. To tackle this challenge, we present our methodologies for the shared task of Multimodal Hate Speech Event Detection at CASE 2023. Our approach combines multimodal deep learning models with text-based tabular features, such as named entities and syntactical features, yielding superior performance compared to existing methods for multimodal hate speech detection. This is evidenced by achieving the first place in both Subtask A and B of the shared task on the final leaderboard, demonstrating the effectiveness of our models in identifying and categorizing hate speech events.

Our study highlights the significance of employing multimodal techniques to detect and combat the dissemination of hate speech and propaganda within text-embedded images. By developing more robust and accurate automated systems for hate speech detection, our findings will contribute to mitigating the negative impact of such content during conflicts and beyond.

\bibliographystyle{acl_natbib}
\bibliography{bibliography}

\end{document}